\documentclass{article}

\usepackage{PRIMEarxiv}

\usepackage[utf8]{inputenc} 
\usepackage[T1]{fontenc}    
\usepackage{hyperref}       
\usepackage{url}            
\usepackage{booktabs}       
\usepackage{amsfonts}       
\usepackage{nicefrac}       
\usepackage{microtype}      
\usepackage{lipsum}
\usepackage{fancyhdr}       
\usepackage{graphicx}       
\usepackage{algorithm}
\usepackage{algpseudocode}
\graphicspath{{media/}}     

\title{SOLIS - The MLOps journey from data acquisition to actionable insights
}

\author{
  Razvan Ciobanu, Alexandru Purdila, Laurentiu Piciu \& Andrei Damian \\
  Lummetry.AI \\
  Bucharest\\
  \texttt{\{razvan, alex, laurentiu, andrei\}@lummetry.ai} \\
}

\begin{document}
\maketitle

\begin{abstract}
Machine Learning operations is unarguably a very important and also one of the hottest topics in Artificial Intelligence lately. Being able to define very clear hypotheses for real-life problems that can be addressed by machine learning models, collecting and curating large amounts of data for model training and validation followed by model architecture search and optimization, then finally presenting the results fits very well the scenario of Data Science experiments. However, this approach does not supply the needed procedures and pipelines to deploy machine learning capabilities in real production-grade systems. Automating live configuration mechanisms, on the fly adapting to live or offline data capture and consumption procedures, serving multiple models in parallel either on edge or cloud architectures, addressing specific limitations of GPU memory or compute power, post-processing inference or prediction results, and serving those either as APIs or with IoT based communication stacks, all these in the same end-to-end pipeline are the real challenges that we try to address in this paper. This paper presents a unified deployment pipeline and a freedom-to-operate approach that supports all the above requirements using the known cross-platform tensor frameworks and script language engines.
\end{abstract}


\section{Introduction}
Since the first coining of the Deep Learning term following the first “big win” of neural approaches marked by AlexNet\cite{krizhevsky2012imagenet}, followed by the wide adoption of GPU-based tensor optimization frameworks such as Tensorflow\cite{abadi2016tensorflow}  and Pytorch\cite{paszke2019pytorch} a new era of machine learning started. Important industry vendors proposed generation after generation of model inference and prediction serving approaches, yet most focused on SaaS and PaaS approaches. Nevertheless, many individual aspects, such as data pipeline preparation and post-processing of model results, remained mostly unsolved due to many options. Besides the cloud computing solutions proposed by the big players on the market, various tools have been developed and released both by the tensorial framework creators - such as Tensorflow-serving\cite{olston2017tensorflow} - as well as by other third parties. However, once again, the problem of data acquisition, preprocessing pipeline, post-inference processing business logic, and final insight delivery remained with the user. Thus, it is now clear that the machine learning development community must strive and find solutions that would help commercial and academic environments migrate from experiment-centric data science to fully functional end-to-end pipelines and applications. MLOps, an area dominated by startups at this particular time, was already becoming a fast-growing market estimated at more than 20 billion USD in 2019 and projected to reach five times this figure by 2025, according to various studies.

\subsection*{Edge vs Cloud}
One of the most common divisions related to the taxonomy of the AI system implementation environments is that defined by the target deployment infrastructure. Thus, there are two main infrastructure options: (i) the general cloud-based approach and (ii) the edge deployment approach. While in the first setting, the serving pipelines and the data pre-processing and business logic reside in Cloud APIs as pipelines of microservices, in the second case, we have most or even all of these components running in local devices - usually embedded devices. While our proposed SOLIS framework focuses on edge "box" devices, we will argue that entire deployments can be quickly and transparently migrated and switched between edge and Cloud-hosted VMs.

\section{Background}

When analyzing the DevOps requirements for production-grade scalable machine learning systems, one can choose various methods for operationalizing the end-to-end pipelines. Currently, we have several options on the market mainly divided into two categories: domain agnostic technologies such as model serving engines that are not biased towards a particular domain and more complex applications that are usually geared towards specific applications - such as time-series prediction or computer vision inference. - usually hosted as Cloud services.

\subsection{Domain agnostic techniques}

One of the most important MLOps pipelines currently on the market used both in academia as well as in actual commercial applications is Kubeflow\cite{bisong2019kubeflow}, an open-source project dedicated to streamlining machine learning pipelines deployment on kubernetes. It focuses on compatibility - running independent and configurable steps, portability - being platform agnostic as well as on scalability - automatic scaling based on load.
It offers deployment solutions for all the machine learning lifecycle steps: data preparation, model training, prediction serving, and service management/monitoring. Although a powerful tool, it requires data science and Kubernetes proficiency. It works by allowing data scientists to develop independent steps. Each step is deployed on its own Docker image in a Docker pod - a single instance of a running process in a Docker environment. These steps are then loosely coupled into a machine learning pipeline. Although each step is easily reusable in new pipelines, Kubeflow offers no inheritance-like mechanism for these steps and components.

A much simple in scope approach is that of TensorFlow Serving\cite{olston2017tensorflow}. TensorFlow Serving is a serving system for machine learning models. It handles the inference and lifetime management of trained models without providing any solution for training, data acquisition, business logic scripting. Its main advantage is that it allows for a low code solution for model deployment in production. It works by deploying one or more “servables” - Tensorflow models, embeddings, vocabularies, etc., on a gRPC or HTTP endpoint. Grouping requests optimize the serving process into batches for joint execution on GPU. As an extra, Tensorflow Serving supports simultaneous serving of different model versions and deployment of new model versions without changing the client code. Nevertheless, this is usually limited to Tensorflow framework models. 

ONNX\cite{shridhar2020interoperating}\cite{bai2019onnx} is an open format designed to represent machine learning models, regardless of what technologies were used to build them. It currently supports conversion from most popular machine learning frameworks like Tensorflow, PyTorch, SciKit-Learn, etc. Its purpose is to provide an interface allowing machine learning applications to be framework agnostic. Current development focuses on model conversion and inference acceleration, but some work has also been done on model training and serving. Being a community-run project, a big part of the development is being done by 3rd party companies, each with its project building upon the standard.

\subsection{Domain oriented engines}

One such example of domain-oriented engines is Microsoft Azure Cognitive Services\cite{del2018introducing} for Computer Vision. This Cloud-based ecosystem provides a set of APIs enabling users to integrate specific functionalities in their systems or applications. The platform currently offers end-to-end functionality for training custom deep learning models, providing user-friendly mechanisms for uploading datasets, training models, and serving models through APIs. However, to the best of our knowledge, at this moment, the platform does not provide you with an out-of-the-box integrated solution capable of real-time data acquisition from multiple sources, inference, post-inference processing with complex business logic, and payload packing for IoT integration. In terms of tensorial framework, Microsoft Azure Cognitive Services allows trained models to be exported in various options like Tensorflow, TensorflowJS, ONNX, or CoreML.

\section{Approach}
As argued in the introduction of this paper, we propose an end-to-end methodology that aims at standardizing a simple yet efficient approach to most of the critical stages of the production-grade machine learning pipelines with a particular focus in the area of “box” systems - i.e., systems that run either in an embedded device or as a Cloud deployed box. This proposal objective is to ensure a maximum level of technological independence for the developers and total freedom to operate while allowing multiple levels of abstraction and ease to deploy new custom features. Particular attention has been paid to aspects such as multi-modal data streams, multi-purpose neural models, multiprocessing and multithreading of various tasks and jobs, plug-and-play post-processing of inference and prediction results, and open-ended delivery of final results. A clear direction of proposed further research and development is that of OmniNet - a multi-purpose, multi-model, multi-stage neural graphs- this direction is briefly presented in the current paper. Last but not least, while our proposed approach has been already implemented and deployed for the computer (deep) vision domain within embedded systems designed for safety and security scenarios, we argue that the presented methods and functionalities can be used cross-domain in other areas such as on-prem or cloud-based predictive analytics.

\subsection{SOLIS Properties}

Several intrinsic properties of our proposed MLOps framework can also be seen as actual objectives: (i) integrating potentially any data source with low-code or no-code while acquiring parallel streams of data; (ii) quick adaptation to any configuration and communication approach from MQTT to HTTP, from flat-files to GraphQL; (iii) machine learning package and tensor framework agnosticism; (iv) low-code to no-code fast creation of business rules for post-processing of inferences and predictions.

\subsubsection{Seamlessly integrate any data source}
Our pipeline stream inception is initiated within a dedicated module for data acquisition that can parallelly gather data from any kind of feed, protocol, or data format, including complex sensors as well as URI-defined devices that are producing data either real-time or not. Data source connections can be established in parallel with self-explanatory and straightforward JSON-based configuration files. As a result, various data formats can be ingested and implemented by employing low-code plugin development in Python. Moreover, multiple inputs streams, including multi-modal ones, can be configured as pre-aggregated streams and thus served downstream to the pipeline as single data packages.

\subsubsection{Versatile communication and configuration for third-parties}

\begin{figure}[H]
    \centering
     \fbox{\includegraphics[width=0.6\textwidth]{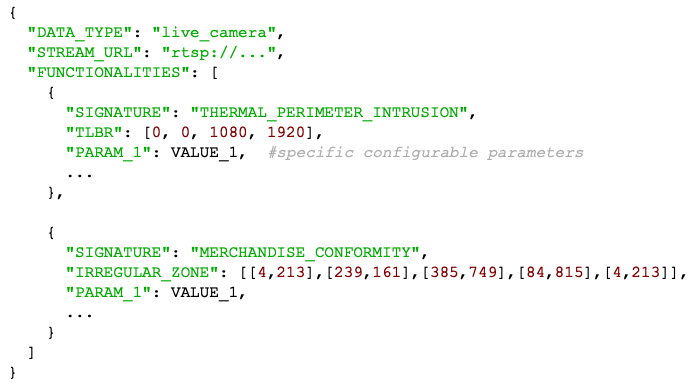}}
     \caption{Configuration file}
     \label{fig:config}
\end{figure}

The \textit{SOLIS} pipeline uses a configurable communication module able to integrate multiple types of protocols such as AMQP, MQTT, HTTP, SQL, GraphQL, with a low-code method for implementing new ones and thus  \textit{SOLIS} can potentially send and receive data to any consumer, regardless of what technology it uses. To further simplify integration with external services, an input-output \textit{formatter} middleware-like module is provided to alter inbound and outbound payloads so that they comply with formats required by each consumer use-case. Each pipeline stage is fully adaptable by using an internal configuration system to provide a customizable framework. This configuration system is split into two sections: the application configuration that handles the main aspects of the pipeline and the configuration of the streams (Figure \ref{fig:config}) that handles data acquisition and functionalities. Since the configuration module allows us to change the system behavior while it runs, specific functionalities can be stopped, started, or changed on the fly.

\subsubsection{Domain and framework agnosticity}

\textit{SOLIS} inference module supports model inference on multiple types of tensor frameworks such as Tensorflow and Pytorch as well as other machine learning libraries like sklearn\cite{pedregosa2011scikit}. 
All the deployment and initial configuration tasks are done based on a set of automatic procedures that handle environment creation, application download, and application configuration. The \textit{SOLIS} pipeline is designed to run on multiple operating systems - Windows, Linux, macOS, etc. and multiple hardware platforms - x86, arm, etc.
As a result of the fully customizable pipeline stages, \textit{SOLIS} is not constrained to any particular domain. As a result, \textit{SOLIS} supports a wide range of end-user business functionalities for multiple areas like Computer Vision or Predictive Analytics.

\subsubsection{Low-code to no-code for developing new business features}

This module is responsible for running specific business rules using all the data provided by the previous pipeline steps, such as model predictions or raw input data. The entire business logic can be implemented in a single Python \textit{plugin}, without knowledge of any technical details regarding the internals of the rest of the pipeline.
What is probably more important is that this business logic they \textit{plugins} can be implemented by engineers with no data science background.
Finally, functionalities can be implemented using a no-code approach, only by specifying  rules to be executed by a particular purpose \textit{plugin}.

\subsection{Pipeline outline}

The main entry point in \textit{SOLIS} is contained in \textit{"main loop"} that orchestrates the execution of all the modules, thus the execution of the entire pipeline. In this section, we will describe the main loop algorithm (Algorithm \ref{alg:mainloop}) with its multi-stage methodology. 

In the first stage of the pipeline, \textit{SOLIS} checks if any new updates have been received since the last check. The communication module is responsible for acquiring any new configuration from external applications, while the configuration module handles the incoming configuration messages and checks their validity. In the second stage, \textit{SOLIS} updates its internal state using the updates received from the first stage. Thus \textit{SOLIS} can start, stop or update data acquisition threads or business functionalities depending on the type of command received from the external application.

The third stage of the pipeline consists of collecting real-time data from all streams in an asynchronous and parallel fashion. Once the data is collected, it is packed and sent downstream. At the fourth and fifth stage of the pipeline, \textit{SOLIS} evaluates which parallel inference and prediction \textit{serving processes} are required while fully managing the process and memory allocation and deallocation. A serving process is defined as an end-to-end inference or prediction pipeline able to run either as an external process or as an internal sequential execution thread. Further downstream, at the sixth stage, the pipeline handles the business features execution based on the upstream raw initial data and the serving process results provided by the fifth stage.

In the last stage of the pipeline, \textit{SOLIS} prepares and sends all the payloads using the communication module. This allows \textit{SOLIS} to repeat asynchronously and in parallel the entire loop while larger payloads are still being sent over.

\begin{algorithm}[H]
\caption{\textit{SOLIS} box "main loop"}\label{alg:mainloop}
\textbf{Input:} $cfg\_steams$
\begin{algorithmic}
    \While{$True$}
    \State $updates \gets$ \textbf{call} receive updates from external application; \vspace{1mm}
    \State $data \gets$ \textbf{async threaded calls} collect data from all streams $(cfg\_streams)$; \vspace{1mm}
    \State $cfg\_streams,\: business\_feats \gets$ \textbf{call} update box internal state $(cfg\_streams,\: updates)$; \vspace{1mm}
    \State $models \gets$ \textbf{call} get business features models $(business\_feats)$; \vspace{1mm}
    \State $inferences \gets$ \textbf{parallel calls} inference $(models, \: data)$; \vspace{1mm}
    \State $payloads \gets$ \textbf{threaded calls} execute $(business\_feats,\: data,\: inferences)$; \vspace{1mm}
    \State \textbf{async threaded calls} send $(payloads)$; \vspace{1mm}
    \EndWhile
\end{algorithmic}
\end{algorithm}

Our past experiences correlated with the work of other practitioners in the field showed us that no machine learning model is performing at its peak across any kind of hardware environment and software conditions. That is why we decided to fine-tune our models based on each specific use case. To obtain this kind of data, a module was designed and implemented in our system with the primary purpose of collecting specific data at regular time intervals or when particular triggers are fired. The collected data is later sent over our model training and fine-tuning pipelines to improve our current performance of machine learning models.

\subsection{The plugin approach}
One essential property of the SOLIS framework is the ability to quickly deploy low-code plugin features at any point in the end-to-end pipeline whenever we discuss adapting to a new custom data source or a complex business logic that must post-process the predictions.
Each module in the \textit{SOLIS} pipeline - i.e., data acquisition, communication, inference, business features, and post-processing - is designed to support no data science skills, low-code, dynamic, and collaborative work approaches within development teams that integrate our system in various applications. One of the most important things we aimed to achieve is \textbf{\textit{no data science or MLOps proficiency required}} in integrators' teams for further development. Based on the system configurations, each module manager initializes the required custom plugins making \textit{SOLIS} a versatile engine that can be integrated in almost any larger application.

Adding new features without compromising usability is a complex and often tedious task for any software development project and even more so in machine-learning-related ones. The \textit{plugin} approach that we have developed enables almost effortless integration of \textit{SOLIS}, even when custom new requirements are yet to be satisfied. Each plugin template of the plugin ecosystem within each \textit{SOLIS} module is well documented and defines very clear methods that should be implemented. The \textit{data acquisition} step can be configured to use any live, non-live, sequential, non-sequential, structured, or unstructured data stream. Multi-modal streams such as streams combining sensor structured data with unstructured video feed can be created, and meta-streams that re-combine multiple input streams into one flow. Regarding the \textit{serving processes} module, it is worth mentioning that it supports any machine learning / deep learning framework. Thus, any in-house or open-source model can be employed.

Having a system that is capable of communicating with the external \textit{world} - i.e., its overall ecosystem - is a must, whether the system runs inside a VPN or it has access over the internet. The \textit{communication} between \textit{SOLIS} any payload consuming applications within proposed ecosystems can be done via MQTT\cite{mqtt}, AMQP\cite{amqp}, HTTP\cite{RFC2616}, sockets or any other protocol that enables message exchange and can be quickly wrapped in a low-code Python \textit{plugin}. By default \textit{SOLIS} already provides MQTT, AMQP \textit{communication plugins}. To make the integration more facile, \textit{SOLIS} does not impose a certain structure of the payloads that are sent by the communication module. Integrating with 3rd parties that already have strict communication protocols and payload definitions is possible through the use of a separate module that is in charge of \textit{post-processing} the system outputs exactly formatted as required by the external application. 

\textbf{\textit{Eyes on the end-client}} - Employing this low code plugin-based method throughout the entire MLOps pipeline empowers development teams to address potentially any kind of end-to-end real-life use-case.

\subsection{Inference}

\subsubsection{OmniNet approach}
The main objective of our proposed OmniNet DAG deployment architecture is to have multiple task-oriented neural models cooperate and similarly integrate with each other to that of hydra networks\cite{mullapudi2018hydranets}. Nevertheless, while the hydra networks usually rely on a backbone feature extractor in our approach, we propose an arbitrary number of backbone graphs that can provide features for subsequent graphs. Complex architectures such as EfficientNet\cite{tan2019efficientnet} backbone running on video volumes (4D) followed by seq-2-seq custom analyzers parallelized with frame-by-frame second-stage classification directed acyclic graphs can be configured both for end-to-end training but more importantly for efficient inference. Other essential features of our proposed OmniNet approach is (i) employing multi-stage graphs fully trainable while using early-stage graphs as “frozen” graphs in order to train later stage ones (ii) fully parallelizable operations are directly optimized on GPU while (iii) keeping low memory footprint both in RAM and VRAM.

\subsubsection{In-process vs parallel multi-process inference and memory management}
A certain challenge when dealing with multiple concurrent directed acyclic graphs (DAG) on GPU is optimizing the inference speed for all the DAGs employed by the business features. Running all the DAGs in the same process imposes sequential running and thus the total inference time at one step is the sum of each individual DAG inference time $T_I = T_{DAG_1} + T_{DAG_2} + ... + T_{DAG_N}$. Therefore, in order to optimize $T_I$, \textit{SOLIS} uses an internally developed parallel multi-process inference execution mechanism with GPU memory management that assures $T_I$ to be $T_I = \max (T_{DAG_1}, T_{DAG_2}, T_{DAG_N}) + \varepsilon$.

Even though speed is a critical parameter in near-real-time machine learning inference-based systems, there is another critical challenge when it comes to concurrency on GPU, and that is \textit{error contention}. Two significant issues emerged from our studies with production-grade ML system, more specifically how to assure that all graphs are executed but the one that generated an error at a certain point (i) for an out-of-memory (OOM) situation, when the GPU tries to accommodate one more DAGs or (ii) if one graph operations generate unpredictable errors when they are executed on GPU. Our proposed solution for these issues is isolating each graph execution in separate processes, making the internally developed parallel multi-process inference execution mechanism a reliable solution with a high degree of fiability and contention for the encountered challenges. While a multi-threading solution would be more simplistic and easy to implement, eliminating the need for data transfer, the multi-process solution solves the error contention.

This approach opens wide horizons in terms of inference - distributed or not - compute and serving configurability. It is worth mentioning that (i) cross-domain tasks can be addressed on the same box, and (ii) the processes can be spawned not only on the box but also on other devices and/or clusters.

While in most of the experimental as well as production environments we used Tensorflow\cite{abadi2016tensorflow} frozen DAGs or similarly Pytorch\cite{paszke2019pytorch} approach based on TorchScript\cite{devito2019torchscript}, any other tensor framework can be used due to the multi-processing approach with no real impediments on using “competing” tensor frameworks in parallel. Thus, for example, one can define a simple Gaussian model in Numpy for simple sensor structured data anomaly detection, use in the same time a complex set of TorchScript\cite{devito2019torchscript} chained directly in GPU memory for minimization of GPU transfer/offload in multi-stage inference, and finally a set of Tensorflow frozen graph models - all of these running on the same box.

\section{Conclusions}

In this whitepaper, we propose an overview for a fully configurable end-to-end methodology that enables easy deployment and configuration of entire application pipelines powered by machine learning, allowing non data scientists to participate in the development process and delivery of custom features to end consumers. The proposed architecture scales well with various computing capabilities and allows multiple tensor computation frameworks and execution engines for the neural models or non-DAG models deployment. While our work is far from being over, we are confident that this architecture will gradually grow into a potential machine learning operations standard. The OmniNet neural model design principles and architectural details have not been presented in detail within this paper, as further research is still underway and will be presented within a subsequent paper.

\bibliographystyle{unsrt}  
\bibliography{references}

\end{document}